\documentclass[letterpaper]{article} % DO NOT CHANGE THIS
\usepackage{aaai2026}  % DO NOT CHANGE THIS
\usepackage{times}  % DO NOT CHANGE THIS
\usepackage{helvet}  % DO NOT CHANGE THIS
\usepackage{courier}  % DO NOT CHANGE THIS
\usepackage[hyphens]{url}  % DO NOT CHANGE THIS
\usepackage{graphicx} % DO NOT CHANGE THIS
\usepackage{natbib}  % DO NOT CHANGE THIS AND DO NOT ADD ANY OPTIONS TO IT
\usepackage{caption} % DO NOT CHANGE THIS AND DO NOT ADD ANY OPTIONS TO IT
\usepackage{algorithm}
\usepackage{algorithmic}
\usepackage[utf8]{inputenc}
\usepackage[T1]{fontenc}
\usepackage{mathtools}
\usepackage{amsmath}
\usepackage{amsfonts}
\usepackage{nicefrac}
\usepackage{soul}
\usepackage{wasysym}
\usepackage{enumitem}
\usepackage{multirow}
\usepackage{booktabs}
\usepackage{comment}
\usepackage{microtype}
\usepackage{xcolor}
\usepackage[table]{xcolor}
\usepackage{cleveref}

\usepackage{newfloat}
\usepackage{listings}
\DeclareCaptionStyle{ruled}{labelfont=normalfont,labelsep=colon,strut=off} % DO NOT CHANGE THIS
\floatstyle{ruled}
\newfloat{listing}{tb}{lst}{}
\floatname{listing}{Listing}
\title{OAD-Promoter: Enhancing Zero-shot VQA using Large Language Models with Object Attribute Description}
\author{
    Quanxing Xu\textsuperscript{\rm 1}, Ling Zhou\textsuperscript{\rm 1}\thanks{Corresponding author.}, Feifei Zhang\textsuperscript{\rm 2}, Jinyu Tian\textsuperscript{\rm 1}, Rubing Huang\textsuperscript{\rm 1,3}
}
\affiliations{
    \textsuperscript{\rm 1}School of Computer Science and Engineering, Macau University of Science and Technology, Macau SAR, China\\

    \textsuperscript{\rm 2}School of Computer Science and Engineering, Tianjin University of Technology, Tianjin, China\\

    \textsuperscript{\rm 3}Macau University of Science and Technology Zhuhai MUST Science and Technology Research Institute, Zhuhai, China\\

    lzhou@must.edu.mo

}

\usepackage{bibentry}
\begin{document}

\maketitle

\begin{abstract}

Large Language Models (LLMs) have become a crucial tool in Visual Question Answering (VQA) for handling knowledge-intensive questions in few-shot or zero-shot scenarios. However, their reliance on massive training datasets often causes them to inherit language biases during the acquisition of knowledge. This limitation imposes two key constraints on existing methods: (1) LLM predictions become less reliable due to bias exploitation, and (2) despite strong knowledge reasoning capabilities, LLMs still struggle with out-of-distribution (OOD) generalization. To address these issues, we propose \underline{\textbf{O}}bject \underline{\textbf{A}}ttribute \underline{\textbf{D}}escription \underline{\textbf{Promoter}} (OAD-Promoter), a novel approach for enhancing LLM-based VQA by mitigating language bias and improving domain-shift robustness. OAD-Promoter comprises three components: the Object-concentrated Example Generation (OEG) module, the Memory Knowledge Assistance (MKA) module, and the OAD Prompt. The OEG module generates global captions and object-concentrated samples, jointly enhancing visual information input to the LLM and mitigating bias through complementary global and regional visual cues. The MKA module assists the LLM in handling OOD samples by retrieving relevant knowledge from stored examples to support questions from unseen domains. Finally, the OAD Prompt integrates the outputs of the preceding modules to optimize LLM inference. Experiments demonstrate that OAD-Promoter significantly improves the performance of LLM-based VQA methods in few-shot or zero-shot settings, achieving new state-of-the-art results. 
\end{abstract}

\section{Introduction}
Different from other VL tasks (e.g., video captioning~\cite{61,62}), Visual Question Answering (VQA)~\cite{1,2}, serves as one of the most important representatives of the multi-modal field, where a natural language question accompanies an image, has attracted considerable attention in recent years due to its requirement for generating accurate natural language responses. 

Nevertheless, language bias remains a significant challenge in this field. For instance, the dominant answer for the question type ``What color \dots bananas?'' in the training data is ``yellow''. Consequently, VQA models may exploit this simple correlation, relying on superficial cues rather than integrating visual information to capture underlying semantics and perform reasoning, which can lead to incorrect answers.

\begin{figure}[t]
	\centering
	\includegraphics[width = 1.0\linewidth]{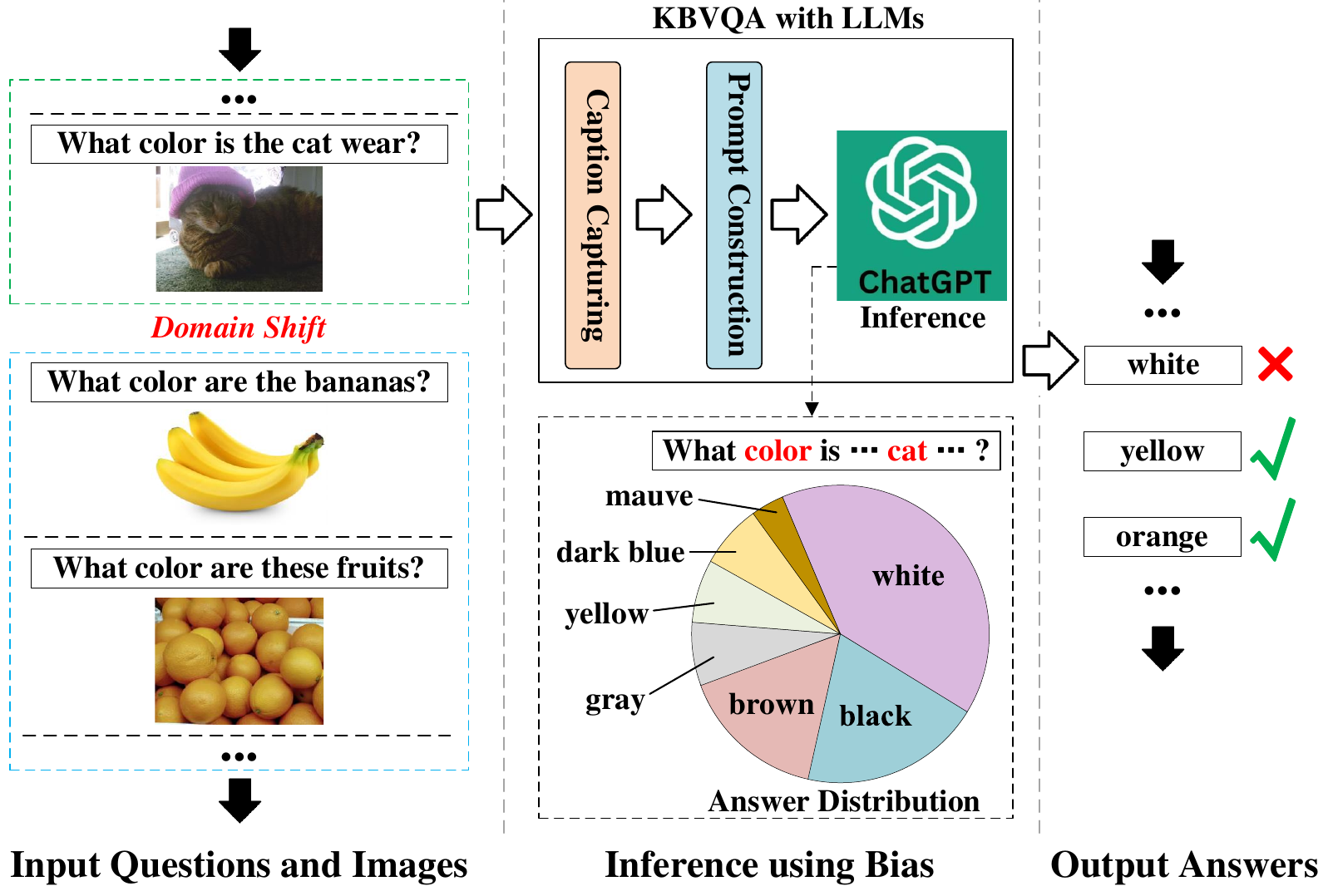}
	% \captionsetup{skip = 1pt}
	\caption{The illustration of the problem in existing LLM-based KBVQA. Like conventional VQA models, LLMs tend to exploit the inside language bias when they conduct inference.  This drawback hampers both their accuracy and domain adaptation capabilities.}
	\label{fig1}
	%\vspace{-10pt}
\end{figure}

In addition, language bias is not only an essential issue in conventional VQA methods, but also a new emerging problem in Knowledge-based VQA (KBVQA)~\cite{6} methods using Large Language Models (LLMs). 
In recent years, LLM-based methods have attained notable advancements in this field. For instance, Yang et al.~\cite{3} proposed a few-shot pipeline for GPT-3~\cite{4}; Guo et al.~\cite{8} successfully realized a zero-shot VQA with OPT~\cite{28}. 
Despite these achievements being dazzling and impressive, few studies have acknowledged the bias problem underlying the pre-trained LLMs. Lately, with the rapid development of pre-trained large models (e.g., Multi-modal Large Language Models (MLLM), Vision-Language Pre-trained models (VLPs), Vision-Language Models (VLMs)), some issues about their drawbacks have come to the surface. The latest works~\cite{9} have implied that the dataset-bias problem can not be underestimated in studies with pre-trained large models. Since these large models were pre-trained on a vast scale of datasets and corpora, they inevitably acquire some spurious correlations instead of the target pattern during the training, which is called shortcut learning~\cite{41}. Therefore, they naturally inherit the bias with the accumulation of learning. 
As illustrated in \cref{fig1}, this drawback has two negative impacts on the LLM-based VQA methods. First, the prediction of LLMs is not reliable enough due to the exploitation of language bias. Second, although LLMs have demonstrated outstanding performance in knowledge reasoning, the out-of-distribution (OOD) circumstance remains a challenging issue, as language bias exacerbates the difficulty for LLMs in adapting to new domains.

The existing LLM-based KBVQA overlooked the combination of the global and regional visual information. Even though LLMs are equipped with tremendous knowledge, it is not easy for them to adapt to new domains seamlessly under a distribution-changing scenario. The invention of a subsidiary memory module to assist LLMs in domain-shift prediction has not been investigated in previous works. 
Inspired by the impressive work~\cite{22,23,42,43}, we recognize that the multi-granularity captions can enhance the integrity and richness of visual information, thereby reducing the impact of language bias; making the memory examples available for the new inference is a refined knowledge supplement for LLMs, consequently improving the reliability of the prediction, especially in domain-shifting situations.

Motivated by the above factors, we posit that
1) More meticulous visual information can mitigate the language bias. 
2) Memory examples assistance can benefit the inference and improve the prediction reliability.
3) A prompt that includes the above two characteristics can constantly promote domain-shift capacity with the inference proceeding.
Accordingly, we propose \underline{\textbf{O}}bject \underline{\textbf{A}}ttribute \underline{\textbf{D}}escription \underline{\textbf{Promoter}} (OAD-Promoter), a LLM-based multiple-module collaborative zero-shot approach designed to overcome inherent language bias in LLMs via a multi-granularity visual description input; and to constantly improve the domain adaptive capability of LLMs via assistant examples support.

The goal of this work is realized by the combination of three components: the Object-concentrated Example Generation (OEG) module, the Memory Knowledge Assistance (MKA) module, and the OAD Prompt. 

Note that the entire OAD-Promoter procedure does not rely on any external knowledge sources or data that need to be retrieved. It is a zero-shot approach.

Our key contributions can be summarized in fourfold:
\begin{itemize}

\item We introduce multi-granularity captions to LLM-based VQA and propose the OEG module, leveraging enhanced visual information to mitigate inherited language biases in LLMs.

\item We design the MKA module to exploit relevant stored object-attribute examples, improving prediction reliability through memory-augmented knowledge support.

\item We develop the OAD Prompt to provide comprehensive visual details and auxiliary examples, enhancing LLM robustness in distribution-changing scenarios during inference.

\item Extensive experiments on \textsc{OKVQA}, \textsc{A-OKVQA}, \textsc{VQAv2}, \textsc{VQA-CP}, and \textsc{GQA-OOD} demonstrate our method's effectiveness and generalizability, establishing new state-of-the-art performance.

\end{itemize}

\begin{figure*}
	\centering
	\includegraphics[width = \textwidth]{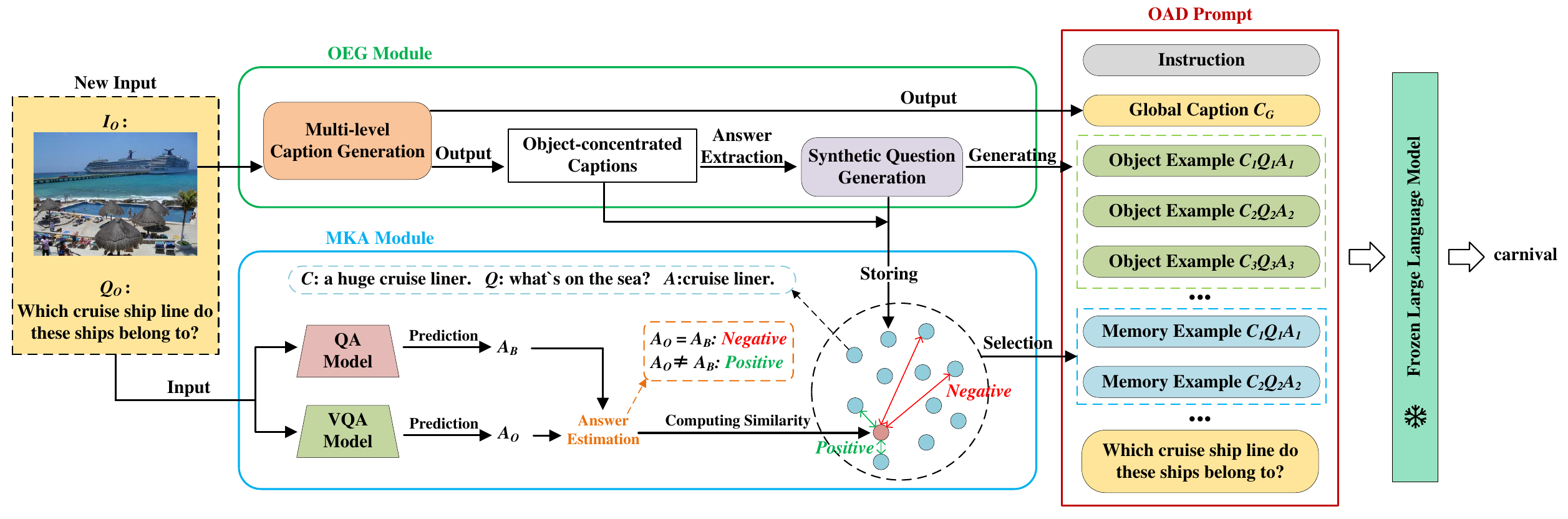}
	\caption{Architecture of OAD-Promoter. It comprises three components: 1) The OEG module (green box) generates a global caption and object-focused samples; 2) The MKA module (blue box) processes novel inputs by leveraging relevant stored examples to assist the LLM; and 3) The OAD Prompt (red box) integrates outputs from the preceding modules and directs the LLM to produce the final answer.
    }
	\label{fig2}
\end{figure*}

\section{Related Work}
\subsection{Debiasing Methods in VQA}
Language bias is one of the most significant issues in VQA, as models are likely to learn easier patterns rather than the target pattern during training. In other words, models tend to remember an unimodal pattern (question-answer correlation) instead of a multi-modal pattern (question-image-answer correlation) during the VQA training~\cite{41}. 
Numerous debiasing methods have been proposed, which can be broadly categorized into five groups: innovative architectures~\cite{20}; methods with an enhanced Language Model (LM); methods with improved visual information~\cite{22,23}; ensemble methods~\cite{21}; data-driven strategies~\cite{37}. Recently, numerous studies have argued that large models still suffer from language bias inherited from their training data.
In this work, we aim to alleviate language bias for VQA using LLMs by enhancing visual descriptions from an object attribute perspective.

\subsection{Knowledge-Based VQA}
There is a trend that LLMs are getting increasingly important in KBVQA. In 2022, Yang et al.~\cite{3} first utilized GPT-3~\cite{4} for KBVQA in a few-shot setting, which they referred to as PICa. Building upon PICa, Shao et al.~\citep{5} further enhanced GPT-3's comprehension of the task by encoding answer heuristics, answer candidates, and answer-aware examples into the prompts. In addition to the few-shot scenarios, Guo et al.~\cite{8} proposed a zero-shot VQA method called Img2LLM, which utilizes the OPT model~\cite{28}. Moreover, some efforts have also focused on improving the prompt's design, such as the Reasoning Question Prompts (RQP) proposed by Lan et al.~\cite{25} for zero-shot VQA, which further promoted the potential of LLMs from a question perspective. Hu et al.~\cite{10} further advanced this study via a novel captioning model that bridges the gap between image and LLMs. Recently, Zhang et al.~\cite{9} focused on improving few-shot LLM-based KBVQA via reducing language bias inherited by LLMs.
In this work, we aim to enhance the domain-shift capability of LLM-based VQA by incorporating practical memory knowledge in a zero-shot setting.

\section{Methodology}
In this section, we present a comprehensive overview of the proposed method. Specifically, we first provide a thorough illustration of the OAD-Promoter architecture, then offer an elaborate explanation of the OEG module and the MKA module, and provide a detailed description of the OAD Prompt.

\subsection{Architecture}
\label{sec: 3.2}
We follow the LLM-based VQA pipeline in previous works~\cite{3,5,8}, the architecture is illustrated in \cref{fig2}, which consists of three components: OEG Module, MKA Module, and OAD Prompt. The OEG Module is responsible for generating a global caption and object-concentrated examples; the MKA module plays a role in assisting LLMs to handle new inputs with the support of relevant stored examples; and the OAD Prompt aims to integrate the output of the former two modules and guide LLMs to predict the output. The proposed method can effectively overcome language bias and improve the LLM's capacity in dealing with OOD samples at the same time. 

To be specific, in the OEG module, motivated by visual-enhancing debiasing methods~\cite{22,23}, we use object-concentrated descriptions to compensate for the absence of fine-grained visual information in the global caption. Thereby, the language bias is mitigated by this visual enhancement approach. Besides, the generated object-concentrated examples are helpful for the LLM in better understanding the detailed visual content of the input image. In the MKA module, to further alleviate language bias and ensure that learned examples remain available in the prompt's architecture, we employ a QA model that outputs a biased answer and a VQA model that conducts an ordinary VQA prediction to pre-assess the language bias of the new input. The computation algorithm depends on the relationship between the outputs (the biased answer and the ordinary answer), and the auxiliary examples are chosen via the similarity computation. In this way, the LLM's exploitation of language bias can be prevented, and its inference on new inputs is strengthened via memory knowledge. In the OAD Prompt, the outputs of the OEG module and the MKA module are integrated into the construction of the prompt. The OAD Prompt's content guides the LLM and predicts the final output. Note that our method's entire procedure does not utilize any external knowledge sources or data that need to be retrieved, making it a pure zero-shot method.     

\begin{figure}[t]
	\centering
	\includegraphics[width = \linewidth]{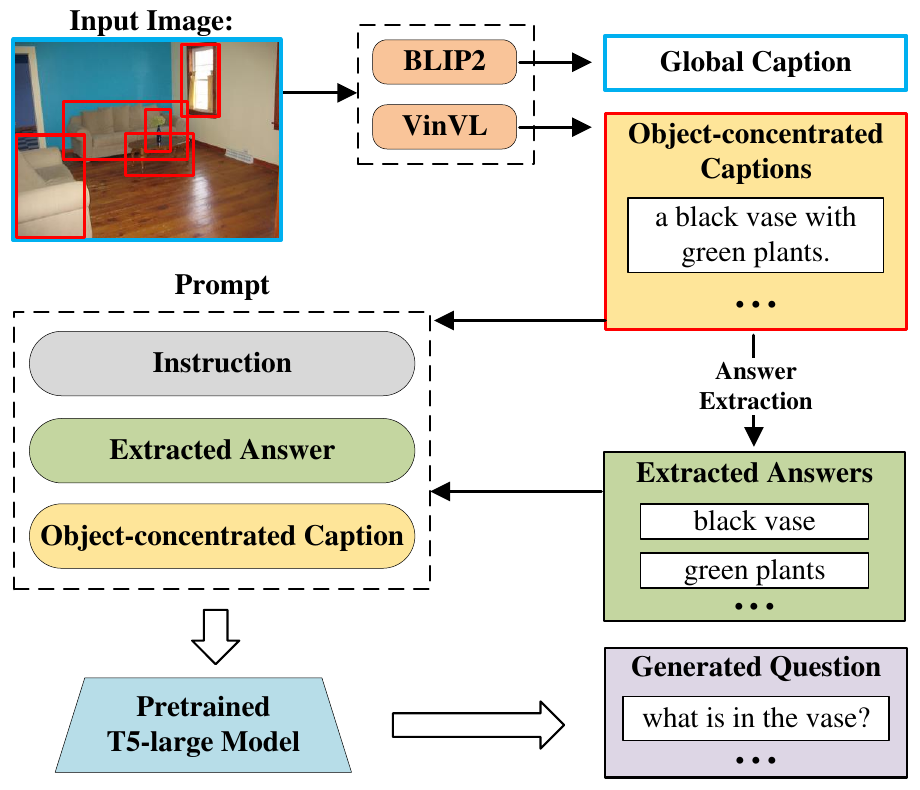}
	\caption{The illustration of the detailed process in the OEG module. Specifically, the BLIP2 and VinVL are used to produce the global caption and object-concentrated captions, respectively. The generated questions are output by a pre-trained T5-large model via prompting.}
	\label{fig3}
\end{figure}

\subsection{OEG Module}
\label{sec: 3.3}
The detailed process in the OEG module is shown in \cref{fig3}. This module comprises two primary generation processes: multi-level caption generation and synthetic question generation. The object-concentrated captions produced by the former process serve as an input to the latter process. 

\textbf{Multi-level Caption Generation.} This process is in charge of generating captions according to the input image. For the input image $I_O$, we utilize a pre-trained BLIP2~\cite{11} and a VinVL detector~\cite{12} to generate a global caption and a couple of regional captions that focus on individual object attributes, respectively.

\textbf{Synthetic Question Generation.} This process is responsible for producing questions according to the generated object-concentrated captions and constitutes complete examples for constructing the prompt. As for an object-concentrated caption, we follow the operation in Img2LLM~\cite{8}, and adopt an effective caption evaluation tool~\cite{13} to extract answers from the caption. Potential answers are extracted in the form of noun phrases, verb phrases, adjective phrases, numbers, as well as boolean-typed words such as ``yes'' and ``no''. After getting potential answers, we utilize a neural question generation method to produce corresponding questions. As illustrated in \cref{fig3}, we form a prompt as: ``[$Instruction$ / $Extracted \quad Answers$ / $Object-concentrated \quad Caption$]'' to guide a pre-trained T5-large model~\cite{14} that finetuned on SQuAD2.0~\cite{15}, MultiRC~\cite{16}, BookQA~\cite{17}, CommonsenseQA~\cite{18} and Social IQa~\cite{19} to produce questions. Utilizing the above question generation approaches, we obtain a collection of synthetic question-answer (QA) pairs for the subsequent operations. 

With the above two processes, an object-concentrated example is formed with the combination of an object-concentrated caption and a synthetic QA pair. Note that each example $E_i$ that contains a caption $C$, a question $Q$, and an answer $A$ is utilized as an element both in the stored memory knowledge of the MKA module and in the prompt construction of the OAD Prompt.

\subsection{MKA Module}
\label{sec: 3.4}
The MKA module is used to assist LLM-based VQA in tackling new inputs by leveraging the memory knowledge from stored object-concentrated examples and supporting the LLM in making more reliable predictions in distribution-changing circumstances. This module comprises two primary processes: answer estimation and similarity computation. The former process aims to estimate the language bias of the new input, while the latter process is responsible for selecting the appropriate stored examples for the subsequent construction of the OAD Prompt.

\textbf{Answer Estimation.} We use a general VQA model~\cite{20} and an off-shift QA model in LMH~\cite{21} in the language bias assessment of the new input. As illustrated in \cref{fig2}, when the input image and question are sent to the MKA module, the VQA model and the QA model both make predictions depending on the given image and question. Then, the answers output by the two models are estimated before the similarity computation starts. Concretely, since the QA model's inference involves no visual information, the result from it is a biased answer, while the result from the VQA model is ordinary. Let $A_B$ and $A_O$ be the output from the QA model and the VQA model, respectively. The selection mode $M$ is defined by the Eq. (\ref{eq1}):
\begin{equation}\label{eq1}
\footnotesize
	M = \begin{cases} 
	Positive, & \text{if } A_O \neq A_B; \\
	Negative, & \text{if } A_O = A_B.
	\end{cases}
\end{equation}
Note that the situation of $A_B$ = $A_O$ means the ordinary VQA model utilizes the language bias during inference, and this implies the LLM can exploit this language bias as well because the scale of the LLM's training is much larger than the ordinary model's training. Therefore, there is a higher probability that the LLM takes advantage of the language bias than a general VQA model does. Hence, in this situation, we adopt a ``Negative'' selection mode that conflicts with the biased result ($A_B$ or $A_O$) in the subsequent similarity computation. On the contrary, if $A_B$ $\neq$ $A_O$, that means there is no sign of the language bias that could be used in the prediction, we choose a ``Positive'' selection mode that supports the ordinary result $A_O$ for enhancing the LLM's inference.

\textbf{Similarity Computation.} Inspired by Prophet~\cite{5}, as for the new input and every stored objective-concentrated example $E_i$, we first extract the fused visual-language feature from the image and the question. Then, the feature $f$ of the new input and the feature $f_i$ of the $E_i$ are the elements in the cosine similarity computation. To be specific, the $f$ is extracted from the aforementioned VQA model in the last process. The visual feature of the $f_i$ comes from the object attribute image of the VinVL, and the language feature comes from the generated question of the pre-trained T5-large model; these two modal features are encoded by the VQA model as well, and $f_i$ is the output. After obtaining the fused features, the cosine similarity is computed by Eq. (\ref{eq2}):
\begin{equation}\label{eq2}
\footnotesize
	E_S = \begin{cases} 
	argTopN\frac{f^T f_j}{\|f\| 2 \|f_j\| 2}, & \text{if } M = Positive; \\
	argBottomN\frac{f^T f_j}{\|f\| 2 \|f_j\| 2}, & \text{if } M = Negative,
	\end{cases}
\end{equation}
where $i \in \{ 1,2, \ldots, K \}$.

The $E_S$ is an index set of the $N$ selected examples in the memory latent space; $K$ is the number of stored examples; $M$ is the selection mode. Note that if $M$ = Positive, we choose the most similar examples in the memory latent space; otherwise, the least similar examples would be selected.

\subsection{OAD Prompt}
\label{sec: 3.5}
The OAD Prompt is responsible for integrating outputs from the former two modules and leading the inference of the LLM. Previous works~\cite{24,25} have proved that the design of prompts is critical in the LLM's inference. Although existing methods have achieved remarkable results in the few-shot or zero-shot settings, they do not account for both global image description and object attribute description. The function of the OAD Prompt includes:
\begin{itemize}
\item Providing more comprehensive and more meticulous image information to LLMs, thereby mitigating language bias by enhanced visual information.

\item Giving similar memory examples for LLMs when it handles the new input.
Thus strengthening the reliability of its prediction with the memory knowledge assistance. 

\item As the inference proceeds, the number of memory examples in the latent space is growing, and the auxiliary memory knowledge is getting more massive and multifarious. Therefore, the domain-shift capability of LLMs is facilitated constantly.

\end{itemize}  

Different from the prompt style in previous works~\cite{3,5,8,25}, the proposed OAD Prompt considers both the global caption and object attribute description of the input image, which is typically overlooked by existing LLM-based VQA methods. The architecture of the OAD Prompt is illustrated in \cref{fig2}, which comprises four parts: the instruction $I$; the global caption $C_G$; the examples $E$; and the inputted question $Q_O$. 
Note that $E$ contains two kinds of examples: object-concentrated examples $E_O$ that are derived from the input image and memory examples $E_S$ that are selected from the memory latent space, and each example can be formulated as $(C, Q, A)$, which represents a triple of a caption, a question, and an answer.

\section{Experiments}

\subsection{Datasets and Metrics}
We evaluate the proposed method on three benchmark datasets: VQAv2~\citep{2}, OKVQA~\citep{6}, and A-OKVQA~\citep{7}, where the image content is not sufficient to answer the questions. Thus, models have to perform reasoning based on the perception of even common sense to answer. Specifically, OKVQA is a commonly used dataset for KBVQA, which contains 9K and 5K image-question pairs for training and testing, respectively. A-OKVQA, an augmented successor of OKVQA, and its questions are more challenging, containing 17K, 1K, and 7K image-question pairs for training, validation, and testing, respectively. Moreover, we utilize VQA-CP~\cite{29} and GQA-OOD~\cite{30} to test the generalization capabilities of the models on unseen data. VQA-CP evaluates the performance of the models on data beyond what they were trained on. Meanwhile, GQA-OOD consists of separate head and tail sets, which are designed for the models to conduct reasoning in unfamiliar scenarios. We use soft accuracy~\citep{2} as the metric in our experiments.

\subsection{Implementation Details}
As for the general VQA model, we utilize a lightweight yet effective UpDn~\cite{20} for encoding fused features and answer estimation. Similar to previous works~\cite{5,26}, we further enhance the model's capabilities by implementing the transfer learning paradigm. Initially, the model undergoes pre-training on the VQAv2 dataset and the Visual Genome dataset~\cite{27}. To avoid data pollution, we eliminate samples from the pre-training dataset if their images are utilized in the testing split of OKVQA. Subsequently, the pre-trained model is further finetuned on the training split of OKVQA to acquire our final VQA model. Regarding the prompt's structure, since the memory module is empty at the beginning, the initial design of the OAD prompt is formed as follows: ``[$I$ / $C_G$ / $E_O$ / $Q_O$]''. And this design turns into ``[$I$ / $C_G$ / $E_O$ / $E_S$ / $Q_O$]'' when the next sample inputted. As for LLMs, we adopt GPT-3~\cite{4} and OPT~\cite{28} as frozen LLMs in our main experiments for a fair comparison with previous methods (i.e., PICa, Prophet, Img2LLM, PromptCap, GRACE). Moreover, we also use other LLMs in the experiment to validate the generalization ability of our method.

\begin{table*}[!t]
	\centering
	\scriptsize
	\setlength\tabcolsep{4.8mm}
	\begin{tabular}{lc|cc|cc|c}
	\toprule[1.1pt]
	\multirow{2}[2]{*}{\textbf{Method}} & \multirow{2}[2]{*}{\textbf{Few/Zero-shot}} & \multicolumn{2}{c|}{\textsc{\textbf{VQAv2}}} & \multicolumn{2}{c|}{\textsc{\textbf{A-OKVQA}}} & \textsc{\textbf{OKVQA}} \\
	\cmidrule(lr){3-4} \cmidrule(lr){5-6} \cmidrule(lr){7-7}
	&  & \textbf{val} & \textbf{test} & \textbf{val} & \textbf{test} & \textbf{test} \\
	\midrule
	\multicolumn{7}{c}{\textit{Methods with Large-scale Multi-modal Pretraining}} \\
	\midrule
	VL-T5~\cite{33}  & Zero-shot & 13.50 & 14.13 & - & - & 5.73  \\
    Frozen~\cite{32}  & Zero-shot & 29.43 & 29.55 & - & - & 5.90  \\
    Flamingo-3B~\cite{31}  & Zero-shot & 48.24 & 49.18 & - & - & 41.17  \\
    Flamingo-9B~\cite{31}  & Zero-shot & 50.77 & 51.80 & - & - & 44.64  \\
    Flamingo-80B~\cite{31}  & Zero-shot & 56.08 & 56.21 & - & - & 50.57  \\
    FewVLM-base~\cite{24}  & Zero-shot & 43.28 & 43.39 & - & - & 11.52  \\
    FewVLM-large~\cite{24}  & Zero-shot & 47.68 & 47.73 & - & - & 16.50  \\
    VLKD-ViT-B/16~\cite{34}  & Zero-shot & 38.60 & 39.69 & - & - & 10.50  \\
    VLKD-ViT-L/14~\cite{34}  & Zero-shot & 42.55 & 44.48 & - & - & 13.24  \\
    \midrule
    \multicolumn{7}{c}{\textit{Methods with Frozen Large Language Models}} \\
    \midrule
    {PICa~\cite{3} \textit{w/} GPT-3}  & Few-shot & 56.10 & 56.12 & 47.08 & 47.64 & 48.01  \\
    {REVIVE~\cite{36} \textit{w/} GPT-3}  & Few-shot & 57.11 & 57.69 & 57.33 & 57.05 & 58.03  \\
    {KAT~\cite{35} \textit{w/} GPT-3}  & Few-shot & 56.30 & 56.48 & 55.38 & 55.12 & 54.41  \\
    {Prophet~\cite{5}\textit{w/} GPT-3}  & Few-shot & 58.37 & 59.22 & \textbf{59.27} & \textbf{57.30} & \textbf{61.08}  \\
    {PromptCap~\cite{10} \textit{w/} GPT-3}  & Few-shot & 58.18 & 58.77 & 58.86 & 57.28 & 60.44  \\
    {GRACE~\cite{9} \textit{w/} GPT-3}  & Few-shot & 58.07 & 58.45 & 58.61 & 57.15 & 60.29  \\
    \rowcolor{gray!20}
    {OAD-Promoter \textit{w/} GPT-3}   & Few-shot & 57.96 & 58.42 & 58.50 & 56.99 & 60.04  \\
    \midrule
    {PICa~\cite{3} \textit{w/} GPT-3}  & Zero-shot & 28.67 & 29.30 & 23.74 & 26.88 & 17.63  \\
    {PICa~\cite{3} \textit{w/} GPT-3 + RQP~\cite{25}}  & Zero-shot & 28.70 & 29.34 & 28.92 & 27.73 & 20.27  \\
    {Img2LLM~\cite{8} \textit{w/} GPT-3}  & Zero-shot & 58.69 & 59.22 & 38.88 & 43.39 & 42.80  \\
    {Img2LLM~\cite{8} \textit{w/} GPT-3 + RQP~\cite{25}}  & Zero-shot & 58.96 & 59.35 & 42.88 & \underline{43.61} & 45.57  \\
    {Img2LLM~\cite{8} \textit{w/} OPT}  & Zero-shot & 60.58 & 61.83 & 42.90 & 40.69 & 45.58  \\
    {Img2LLM~\cite{8} \textit{w/} OPT + RQP~\cite{25}}  & Zero-shot & 60.60 & 61.85 & 42.83 & 40.64 & 45.52  \\
    \rowcolor{gray!20}
    {OAD-Promoter \textit{w/} OPT}   & Zero-shot & 60.62 & 61.93 & 43.03 & 40.68 & 45.58  \\
    \rowcolor{gray!20}
    {OAD-Promoter \textit{w/} GPT-3}   & Zero-shot & \underline{\textbf{60.64}} & \underline{\textbf{61.98}} & \underline{43.09} & 41.71 & \underline{45.61}  \\
	\bottomrule[1.1pt]
	\end{tabular}%
    \caption{Performance comparison on \textsc{VQAv2}, \textsc{A-OKVQA}, and \textsc{OKVQA test set}. The bold fonts indicate the best results in the entire table, and the underlined fonts denote the best results among methods with frozen LLMs in the zero-shot setting. Note that we use GPT-3 in this experiment for the sake of a fair comparison.}
	\label{tab1}%
\end{table*}

\subsection{Quantitative Results}
We conduct comparative experiments with various KBVQA methods and methods that need large-scale vision-language pretraining. All methods used in experiments follow their official instructions.
The main quantitative results are summarized in \cref{tab1}. \textbf{Compared with the methods with large-scale multi-modal pretraining,} OAD-Promoter shows remarkable performance that is superior to all of them. \textbf{Compared with the methods with frozen LLMs,} OAD-Promoter yields competitive results both in few-shot and zero-shot settings. Especially, a new state-of-the-art result was attained on the \textsc{VQAv2} dataset under a zero-shot scenario.

We also compare OAD-Promoter with the latest method (GRACE~\cite{9}) using different LLMs (i.e., LLaVA-1.5~\cite{44}, LLaMA2~\cite{45}, and GPT-4~\cite{4}) in few-shot scenario and report results in \cref{tab2}. Although our results are inferior to those of LLaVA-1.5 and LLaMA2, our method achieves the best accuracy on \textsc{GQA-OOD} with GPT-4, which suggests that OAD-Promoter can demonstrate a better domain-shift capacity with the more advanced frozen LLM.  

For further validation of the zero-shot ability and generalization capability of our methods, we use different sizes of diverse LLMs (i.e., GPT-3~\cite{4}, OPT~\cite{28}, BLOOM~\cite{46}, GPT-Neo~\cite{48}, GPT-J~\cite{47}) in zero-shot evaluation on \textsc{OKVQA test set} with other zero-shot KBVQA methods, the results are presented in \cref{tab3}. These results highlight that the proposed method can integrate various existing LLMs, which proves the generalization competence of our method.  

Moreover, as illustrated in \cref{tab4}, to validate the effectiveness of the proposed method compared with the debiasing integration, we conduct a few-shot comparison experiment to compare three methods with the debiasing methods~\cite{21,37} that can be integrated into the LLM-based pipeline. From the accuracy of \textsc{OKVQA}, we can see the result drops with the integration of debiasing methods. This phenomenon demonstrates that LLM-based methods tend to exploit language bias when addressing knowledge-intensive questions, and it suggests that there is indeed a lack of reliability in the answer output by LLMs as well.

\begin{table}[t]
	\centering
	\scriptsize
	\setlength{\tabcolsep}{1pt}
	\setlength\tabcolsep{2.5mm}
	\begin{tabular}{l|cc|cc}
	\toprule[1.1pt]
	\multirow{2}[2]{*}{\textbf{LLM}} & \multicolumn{2}{c|}{\textbf{GRACE}} & \multicolumn{2}{c}{\cellcolor{gray!20}\textbf{OAD-Promoter}} \\
	\cmidrule(lr){2-3} \cmidrule(lr){4-5} 
	& \textsc{\textbf{VQA-CP}} & \textsc{\textbf{GQA-OOD}} & \textsc{\textbf{VQA-CP}} & \textsc{\textbf{GQA-OOD}} \\
	\midrule
	LLaVA-1.5 & \underline{\textbf{54.08}} & \underline{\textbf{48.96}} &  53.94 &  48.93 \\
	LLaMA2 &\underline{\textbf{57.32}} & \underline{\textbf{50.20}} &  55.89 & 49.67 \\
	GPT-4 & \underline{\textbf{57.61}} &  50.19 &  55.93 & \underline{\textbf{50.21}} \\
	\bottomrule[1.1pt]
	\end{tabular}
    \caption{Few-shot performance comparison on \textsc{VQA-CP} and \textsc{GQA-OOD} with different LLMs. The best results are highlighted in bold and underlined.}
	\label{tab2}
    \vspace{-4pt}
\end{table}

\begin{figure}[h]
	\centering
	\includegraphics[width = \linewidth]{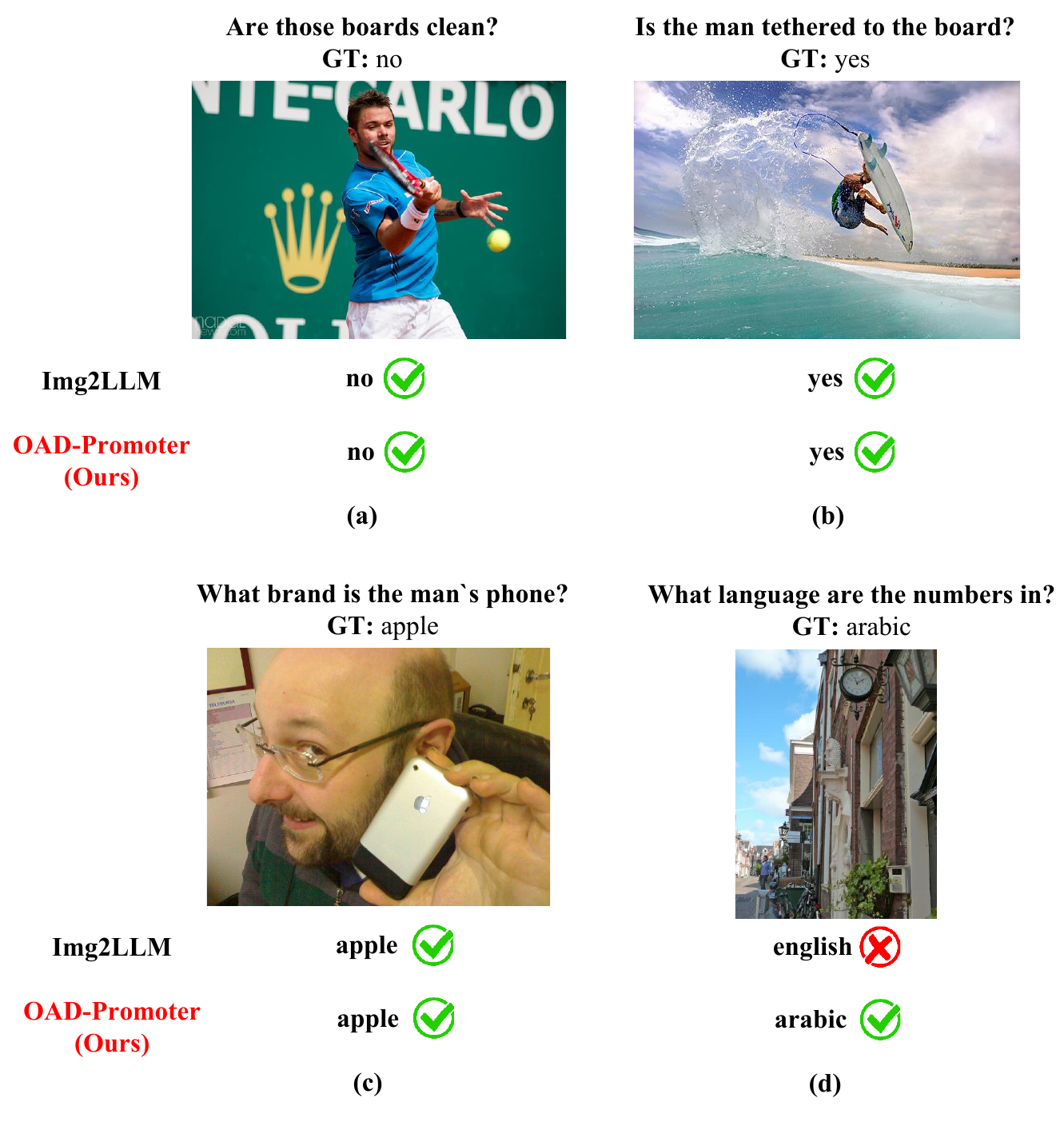}
	% \captionsetup{skip = 1pt}
	\caption{Qualitative analysis of the proposed method. Four cases from distinct domains are displayed.}
	\label{fig4}
\end{figure}

\subsection{Qualitative Analysis}
The qualitative analysis is shown in \cref{fig4}. The representative zero-shot KBVQA method (Img2LLM~\cite{8}) is compared with the OAD-Prompter, and four cases are shown here. Note that cases (a-b) are from close domains concerned with sport and athletics; however, case (c) comes from a different domain that is about electronic devices or technology brands, and case (d) derives from a highly distinct domain in language. These cases are sent to the two methods in alphabetical order. Our method achieves 100\% correctness in four cases, although these cases involve three different domains. The Img2LLM attains 75\% accuracy, but it fails in case (d). 
In addition, we reverse the input order of these four cases and observe the results of the two methods. As a result, our method still maintains 100\% correctness, and Img2LLM also achieves all right predictions. This interesting situation uncovers that the input order also has an influence on the LLM's inference when it deals with samples from multiple different domains
The experiment confirms that our method exhibits better adaptation capability in a distribution-changing environment.  

\begin{table}[t]
  \centering
  \scriptsize
  \setlength\tabcolsep{6mm}
  %\captionsetup{skip=1pt}
    \begin{tabular}{l|cc}
    \toprule[1.1pt]
    \textbf{Method} & \textbf{LLM Size} & \textbf{OKVQA} \\
    \midrule
    Frozen & 7B    & 5.90  \\
    {PICa \textit{w/} GPT-3} & 175B  & 17.63  \\
    {Img2LLM \textit{w/} OPT} & 6.7B  & 38.20  \\
    {Img2LLM \textit{w/} OPT} & 30B   & 41.82  \\
    {Img2LLM \textit{w/} OPT} & 175B  & 45.58  \\
    \midrule
    \rowcolor{gray!20}
    {OAD-Promoter \textit{w/} BLOOM} & 7.1B  & 33.77  \\
    \rowcolor{gray!20}
    {OAD-Promoter \textit{w/} OPT} & 6.7B  & 36.18  \\
    \rowcolor{gray!20}
    {OAD-Promoter \textit{w/} OPT} & 30B   & 40.46  \\
    \rowcolor{gray!20}
    {OAD-Promoter \textit{w/} OPT} & 175B  & 45.58  \\
    \rowcolor{gray!20}
    {OAD-Promoter \textit{w/} GPT-Neo} & 2.7B  & 33.41  \\
    \rowcolor{gray!20}
    {OAD-Promoter \textit{w/} GPT-J} & 6B    & 38.89  \\
    \rowcolor{gray!20}
    {OAD-Promoter \textit{w/} GPT-3} & 175B  & 45.61  \\
    \bottomrule
    \end{tabular}
  \caption{Zero-shot evaluation of the OAD-Promoter based on different sizes of diverse LLMs.}
  \label{tab3}
\end{table}

\begin{table}[!t]
  \centering
  \scriptsize
  \setlength\tabcolsep{6.6mm}
    \begin{tabular}{l|cc}
    \toprule[1.1pt]
    \textbf{Method} & \textsc{\textbf{OKVQA}} & \textsc{\textbf{VQA-CP}} \\
    \midrule
    PICa & 48.01 & 41.90 \\
    PICa + LMH & 47.91 $\downarrow$  & 43.13  \\
    PICa + LMH + CSS & 46.06 $\downarrow$  & 46.87  \\
    \midrule
    Prophet & \textbf{61.11} & 53.41 \\
    Prophet + LMH & 59.96 $\downarrow$  & 54.33  \\
    Prophet + LMH + CSS & 57.28 $\downarrow$  & 55.31  \\
    \midrule
    GRACE & 60.32 & 57.35 \\
    GRACE + LMH & 59.92 $\downarrow$  & 58.72  \\
    GRACE + LMH + CSS & 60.15 $\downarrow$  & \textbf{61.37}  \\
    \midrule
    \rowcolor{gray!20}
    OAD-Promoter & 60.04  & 56.47  \\
    \bottomrule
    \end{tabular}
  \caption{Few-shot evaluation compared with KBVQA methods with debiasing strategies LHM~\cite{21} and CSS~\cite{37}.}
  \label{tab4}
\end{table}

\begin{table}[!t]
	\centering
	\scriptsize
	\setlength\tabcolsep{5.8mm}
	\begin{tabular}{lcc|c}
	\toprule[1.1pt]
	\textbf{Method} & \textbf{OEG} & \textbf{MKA} & \textsc{\textbf{OKVQA}} \\
	\midrule
    \multicolumn{4}{c}{\textit{Few-shot setting}} \\
	\midrule
	OAD-Promoter & \scalebox{2}{$\circ$} & \scalebox{2}{$\circ$} & 47.33 \\
	OAD-Promoter & \scalebox{2}{$\bullet$} & \scalebox{2}{$\circ$} & 54.68 \\
	OAD-Promoter & \scalebox{2}{$\circ$} & \scalebox{2}{$\bullet$} & 48.95  \\
    \rowcolor{gray!20}
    OAD-Promoter & \scalebox{2}{$\bullet$} & \scalebox{2}{$\bullet$} & 60.04 \\
	\midrule
    \multicolumn{4}{c}{\textit{Zero-shot setting}} \\
	\midrule
	OAD-Promoter & \scalebox{2}{$\circ$} & \scalebox{2}{$\circ$} & 42.50 \\
	OAD-Promoter & \scalebox{2}{$\bullet$} & \scalebox{2}{$\circ$} & 44.26 \\
	OAD-Promoter & \scalebox{2}{$\circ$} & \scalebox{2}{$\bullet$} & 43.64  \\
    \rowcolor{gray!20}
    OAD-Promoter & \scalebox{2}{$\bullet$} & \scalebox{2}{$\bullet$} & 45.61 \\
	\bottomrule[1.1pt]
	\end{tabular}
    \caption{Ablation study on the proposed OEG and MKA modules on \textsc{OKVQA test set}.}
	\label{tab5}
\end{table}

\begin{table}[!t]
	\centering
	\scriptsize
	\setlength\tabcolsep{4.3mm}
	% \captionsetup{skip = 1pt}
	\begin{tabular}{lcc|c|c}
	\toprule[1.1pt]
	\textbf{Method} & \textbf{OEG} & \textbf{MKA} & \textbf{K} & \textsc{\textbf{OKVQA}} \\
	\midrule
	OAD-Promoter & \scalebox{2}{$\circ$} & \scalebox{2}{$\bullet$} & 0 & 43.64  \\
    OAD-Promoter & \scalebox{2}{$\circ$} & \scalebox{2}{$\bullet$} & 60 & 43.65  \\
    OAD-Promoter & \scalebox{2}{$\circ$} & \scalebox{2}{$\bullet$} & 200 & 43.92  \\
    OAD-Promoter & \scalebox{2}{$\circ$} & \scalebox{2}{$\bullet$} & 400 & 44.15 \\
	\bottomrule[1.1pt]
	\end{tabular}
    \caption{Ablation study on the effect of example number in the MKA module. Note that we introduce $K$ examples into the MKA module manually at the beginning.}
	\label{tab6}
\end{table}

\begin{table}[!t]
  \centering
  \scriptsize
  \setlength\tabcolsep{14.3mm}
  %\captionsetup{skip=1pt}
    \begin{tabular}{l|c}
    \toprule[1.1pt]
    \textbf{Prompt Design} & \textbf{OKVQA} \\
    \midrule
    CCC-QAQAQA & 44.82 \\
    CQA-CQA-CQA & 45.61 \\
    \bottomrule
    \end{tabular}
  \caption{Ablation study on the OAD Prompt with two prompt designs on \textsc{OKVQA test set} under zero-shot setting.}
  \label{tab7}
\end{table}

\subsection{Ablation Study}
\label{sec: ablation}
To assess the contributions of the proposed OEG and MKA modules, we conduct ablation studies by evaluating our method with and without each module. Note that we use the example integration method in PICa and Img2LLM to substitute the OEG module in the few-shot and zero-shot settings, respectively.
The results in \cref{tab5} demonstrate that both the OEG and MKA modules jointly contribute to performance improvements in our architecture. Moreover, as illustrated in \cref{tab6}, to probe the influence of example number $K$ on the accuracy as the inference proceeds, we manually add examples in a fixed number in the MKA module at the beginning. Note that, unlike the few-shot setting, we provide numerous external examples for the MKA module at the very beginning of the inference. The results reveal that the more examples are provided, the better the performance.

To explore the effect of prompt design on the LLM's inference, we conduct experiments with two designs of the OAD Prompt and report their results in \cref{tab7}. 
The $C$, $Q$, and $A$ stand for context, question, and answer in one example, respectively. Assuming the number of examples is 3, the first design is $CCC-QAQAQA$, the second is $CQA-CQA-CQA$. 
The results illustrated that a complete entirety, which contains a context, a question, and an answer, is more helpful for the OAD Prompt.

\section{Conclusion}
In this work, we propose OAD-Promoter, a novel zero-shot KBVQA method designed to mitigate language bias and strengthen domain-shift robustness in LLMs by providing multi-granularity captions and auxiliary stored examples as enhanced information to LLMs. 
By integrating three components: OEG module, MKA module, and OAD Prompt, OAD-Promoter achieves competitive performance compared to few-shot LLM-based methods and establishes a new state-of-the-art among zero-shot LLM-based approaches. Comprehensive quantitative and qualitative experiments on \textsc{OKVQA}, \textsc{A-OKVQA}, \textsc{VQAv2}, \textsc{VQA-CP}, and \textsc{GQA-OOD} datasets validate its effectiveness. We believe OAD-Promoter can contribute to advancing LLM-based VQA research in zero-shot and real-world applications.

\section{Acknowledgments}
This work was in part supported by the Science and Technology Development Fund of Macau, Macau SAR (Grants No. 0035/2023/ITP1 and 0021/2023/RIA1); the National Natural Science Foundation of China (Grant No. 62376196); and the Tianjin Natural Science Foundation (Grant No. 24JCJQJC00190).

\bibliography{aaai2026}

@InProceedings{1,
  author       = {Stanislaw Antol and
                  Aishwarya Agrawal and
                  Jiasen Lu and
                  Margaret Mitchell and
                  Dhruv Batra and
                  C. Lawrence Zitnick and
                  Devi Parikh},
  title        = {{VQA:} Visual Question Answering},
  booktitle={Proceedings of the {IEEE/CVF} International Conference on Computer Vision},
  pages={2425--2433},
  year={2015}
}

@InProceedings{2,
  author       = {Yash Goyal and
                  Tejas Khot and
                  Aishwarya Agrawal and
                  Douglas Summers{-}Stay and
                  Dhruv Batra and
                  Devi Parikh},
  title        = {Making the {V} in {VQA} Matter: Elevating the Role of Image Understanding
                  in Visual Question Answering},
  booktitle      = {Proceedings of the {IEEE/CVF} Conference on Computer Vision and Pattern Recognition},
  pages={6904--6913},
  year={2017}
}

@InProceedings{3,
  title     = {An empirical study of {GPT-3} for few-shot knowledge-based {VQA}},
  author    = {Zhengyuan Yang and
                  Zhe Gan and
                  Jianfeng Wang and
                  Xiaowei Hu and
                  Yumao Lu and
                  Zicheng Liu and
                  Lijuan Wang},
  booktitle = {Proceedings of the {AAAI} Conference on Artificial Intelligence},
  pages     = {3081--3089},
  year      = {2022}
}

@inproceedings{4,
  author       = {Tom B. Brown and
                  Benjamin Mann and
                  Nick Ryder and
                  Melanie Subbiah and
                  Jared Kaplan and
                  Prafulla Dhariwal and
                  Arvind Neelakantan and
                  Pranav Shyam and
                  Girish Sastry and
                  Amanda Askell and
                  others},
  title        = {Language Models are Few-Shot Learners},
  booktitle    = {Proceedings of the International Conference on Neural Information Processing Systems},
pages = {1877--1901},
  year         = {2020}
}

@InProceedings{5,
  title     = {Prompting large language models with answer heuristics for knowledge-based visual question answering},
  author    = {Zhenwei Shao and Zhou Yu and Meng Wang and Jun Yu},
  booktitle = {Proceedings of the {IEEE/CVF} Conference on Computer Vision and Pattern Recognition},
  pages     = {14974--14983},
  year      = {2023}
}

@inproceedings{6,
  title        = {{OK-VQA:} {A} Visual Question Answering Benchmark Requiring External
                  Knowledge},
  author       = {Kenneth Marino and
                  Mohammad Rastegari and
                  Ali Farhadi and
                  Roozbeh Mottaghi},
  booktitle    = {Proceedings of the {IEEE/CVF} Conference on Computer Vision and Pattern Recognition},
  pages        = {3195--3204},
  year         = {2019}
}

@InProceedings{7,
  title        = {{A-OKVQA:} A benchmark for visual question answering using world knowledge},
  author       = {Dustin Schwenk and
                  Apoorv Khandelwal and
                  Christopher Clark and
                  Kenneth Marino and
                  Roozbeh Mottaghi},
  booktitle    = {Proceedings of the European Conference on Computer Vision},
  pages        = {146--162},
  year         = {2022}
}

@InProceedings{8,
  title     = {From images to textual prompts: Zero-shot visual question answering with frozen large language models},
  author    = {Jiaxian Guo and 
               Junnan Li and 
               Dongxu Li and 
               Anthony Meng Huat Tiong and 
               Boyang Li and 
               Dacheng Tao and 
               Steven Hoi},
  booktitle = {Proceedings of the {IEEE/CVF} Conference on Computer Vision and Pattern Recognition},
  pages     = {10867--10877},
  year      = {2023}
}

@inproceedings{9,
  title        = {{GRACE:} Graph-Based Contextual Debiasing for Fair Visual Question
                  Answering},
  author       = {Yifeng Zhang and
                  Ming Jiang and
                  Qi Zhao},
  booktitle    = {Proceedings of the European Conference on Computer Vision},
  pages        = {176--194},
  year         = {2024}
}

@InProceedings{10,
  author       = {Yushi Hu and
                  Hang Hua and
                  Zhengyuan Yang and
                  Weijia Shi and
                  Noah A. Smith and
                  Jiebo Luo},
  title        = {{PromptCap:} Prompt-Guided Task-Aware Image Captioning},
  booktitle    = {Proceedings of the {IEEE/CVF} International Conference on Computer Vision},
  pages        ={2963--2975},
  year         = {2023}
}

@inproceedings{11,
  author       = {Junnan Li and
                  Dongxu Li and
                  Silvio Savarese and
                  Steven C. H. Hoi},
  title        = {{BLIP-2:} Bootstrapping Language-Image Pre-training with Frozen Image Encoders and Large Language Models},
  booktitle    = {Proceedings of the International Conference on Machine Learning},
  pages        = {19730--19742},
  year         = {2023}
}

@inproceedings{12,
  author       = {Pengchuan Zhang and
                  Xiujun Li and
                  Xiaowei Hu and
                  Jianwei Yang and
                  Lei Zhang and
                  Lijuan Wang and
                  Yejin Choi and
                  Jianfeng Gao},
  title        = {VinVL: Revisiting Visual Representations in Vision-Language Models},
  booktitle    = {Proceedings of the {IEEE/CVF} Conference on Computer Vision and Pattern Recognition},
  pages        = {5579--5588},
  year         = {2021}
}

@inproceedings{13,
  author       = {Hwanhee Lee and
                  Thomas Scialom and
                  Seunghyun Yoon and
                  Franck Dernoncourt and
                  Kyomin Jung},
  title        = {{QACE:} Asking Questions to Evaluate an Image Caption},
  booktitle    = {Findings of the Conference on Empirical Methods in Natural Language Processing},
  pages        = {4631--4638},
  year         = {2021}
}

@article{14,
  title={Exploring the limits of transfer learning with a unified text-to-text transformer},
  author={Colin Raffel and 
          Noam Shazeer and 
          Adam Roberts and 
          Katherine Lee and 
          Sharan Narang and 
          Michael Matena and 
          Yanqi Zhou and 
          Wei Li and 
          Peter J Liu},
  journal={Journal of Machine Learning Research},
  volume={21},
  number={140},
  pages={1--67},
  year={2020}
}

@InProceedings{15,
  author={Pranav Rajpurkar and Robin Jia and Percy Liang},
  title={Know what you don't know: Unanswerable questions for SQuAD},
  booktitle={Proceedings of the Annual Meeting of the Association for Computational Linguistics},
  pages={784--789},
  year={2018}
}

@inproceedings{16,
  author={Daniel Khashabi and Snigdha Chaturvedi and Michael Roth and Shyam Upadhyay and Dan Roth},
  title={Looking beyond the surface: A challenge set for reading comprehension over multiple sentences},
  booktitle={Proceedings of the Conference of the North American Chapter of the Association for Computational Linguistics},
  pages={252--262},
  year={2018}
}

@InProceedings{17,
  author={Todor Mihaylov and Peter Clark and Tushar Khot and Ashish  Sabharwal},
  title={Can a suit of armor conduct electricity? a new dataset for open book question answering},
  booktitle    = {Proceedings of the Conference on Empirical Methods in Natural Language Processing},
  pages={2381--2391},
  year={2018}
}

@InProceedings{18,
  author={Alon Talmor and Jonathan Herzig and Nicholas Lourie and Jonathan Berant},
  title={Commonsense{QA}: A question answering challenge targeting commonsense knowledge},
  booktitle={Proceedings of the Conference of the North American Chapter of the Association for Computational Linguistics},
  pages={4149--4158},
  year={2019}
}

@InProceedings{19,
  author={Maarten Sap and Hannah Rashkin and Derek Chen and Ronan LeBras and Yejin Choi},
  title={Social{IQ}a: Commonsense reasoning about social interactions},
  booktitle    = {Proceedings of the Conference on Empirical Methods in Natural Language Processing},
  pages={4463--4473},
  year={2019}
}

@inproceedings{20,
  title        = {Bottom-Up and Top-Down Attention for Image Captioning and Visual Question
                  Answering},
  author       = {Peter Anderson and
                  Xiaodong He and
                  Chris Buehler and
                  Damien Teney and
                  Mark Johnson and
                  Stephen Gould and
                  Lei Zhang},
  booktitle    = {Proceedings of the {IEEE/CVF} Conference on Computer Vision and Pattern Recognition},
  pages        = {6077--6086},
  year         = {2018}
}

@inproceedings{21,
  title        = {Don't Take the Easy Way Out: Ensemble Based Methods for Avoiding Known
                  Dataset Biases},
  author       = {Christopher Clark and
                  Mark Yatskar and
                  Luke Zettlemoyer},
  booktitle    = {Proceedings of the Conference on Empirical Methods in Natural Language Processing},
  pages        = {4067--4080},
  year         = {2019}
}

@inproceedings{22,
  author       = {Ramprasaath Ramasamy Selvaraju and
                  Stefan Lee and
                  Yilin Shen and
                  Hongxia Jin and
                  Shalini Ghosh and
                  Larry P. Heck and
                  Dhruv Batra and
                  Devi Parikh},
  title        = {Taking a {HINT:} Leveraging Explanations to Make Vision and Language
                  Models More Grounded},
  booktitle    = {Proceedings of the {IEEE/CVF} International Conference on Computer Vision},
  pages        = {2591--2600},
  year         = {2019}
}

@inproceedings{23,
  author       = {Jialin Wu and
                  Raymond J. Mooney},
  title        = {Self-Critical Reasoning for Robust Visual Question Answering},
  booktitle    = {Proceedings of the International Conference on Neural Information Processing Systems},
  pages        = {8604--8614},
  year         = {2019}
}

@inproceedings{24,
  author       = {Woojeong Jin and
                  Yu Cheng and
                  Yelong Shen and
                  Weizhu Chen and
                  Xiang Ren},
  title        = {A Good Prompt Is Worth Millions of Parameters: Low-resource Prompt-based
                  Learning for Vision-Language Models},
  booktitle    = {Proceedings of the Annual Meeting of the Association for Computational Linguistics},
  pages        = {2763--2775},
  year         = {2022}
}

@inProceedings{25,
  title     = {Improving zero-shot visual question answering via large language models with reasoning question prompts},
  author    = {Yunshi Lan and 
               Xiang Li and 
               Xin Liu and 
               Yang Li and 
               Wei Qin and 
               Weining Qian},
  booktitle = {Proceedings of the {ACM} International Conference on Multimedia},
  pages     = {4389--4400},
  year      = {2023}
}

@inproceedings{26,
  author       = {Kenneth Marino and
                  Xinlei Chen and
                  Devi Parikh and
                  Abhinav Gupta and
                  Marcus Rohrbach},
  title        = {{KRISP:} Integrating Implicit and Symbolic Knowledge for Open-Domain
                  Knowledge-Based {VQA}},
  booktitle    = {Proceedings of the {IEEE/CVF} Conference on Computer Vision and Pattern Recognition},
  pages        = {14111--14121},
  year         = {2021}
}

@article{27,
  author       = {Ranjay Krishna and
                  Yuke Zhu and
                  Oliver Groth and
                  Justin Johnson and
                  Kenji Hata and
                  Joshua Kravitz and
                  Stephanie Chen and
                  Yannis Kalantidis and
                  Li{-}Jia Li and
                  David A. Shamma and
                  others},
  title        = {Visual Genome: Connecting Language and Vision Using Crowdsourced Dense
                  Image Annotations},
  journal      = {International Journal of Computer Vision},
  volume       = {123},
  number       = {1},
  pages        = {32--73},
  year         = {2017}
}

@Article{28,
  title   = {{OPT}: Open pre-trained transformer language models},
  author       = {Susan Zhang and
                  Stephen Roller and
                  Naman Goyal and
                  Mikel Artetxe and
                  Moya Chen and
                  Shuohui Chen and
                  Christopher Dewan and
                  Mona T. Diab and
                  Xian Li and
                  Xi Victoria Lin and
                  others},
  journal = {arXiv preprint arXiv:2205.01068},
  year    = {2022},
  note    = {\url{https://doi.org/10.48550/arXiv.2205.01068}}
}

@inproceedings{29,
  author       = {Aishwarya Agrawal and
                  Dhruv Batra and
                  Devi Parikh and
                  Aniruddha Kembhavi},
  title        = {Don't Just Assume; Look and Answer: Overcoming Priors for Visual Question
                  Answering},
  booktitle    = {Proceedings of the {IEEE/CVF} Conference on Computer Vision and Pattern Recognition},
  pages        = {4971--4980},
  year         = {2018}
}

@inproceedings{30,
  title        ={Roses are red, violets are blue... but should VQA expect them to?},
  author       = {Corentin Kervadec and
                  Grigory Antipov and
                  Moez Baccouche and
                  Christian Wolf},
  booktitle    ={Proceedings of the {IEEE/CVF} Conference on Computer Vision and Pattern Recognition},
  pages        ={2776--2785},
  year         ={2021}
}

@inproceedings{31,
  title   = {Flamingo: a visual language model for few-shot learning},
  author       = {Jean{-}Baptiste Alayrac and
                  Jeff Donahue and
                  Pauline Luc and
                  Antoine Miech and
                  Iain Barr and
                  Yana Hasson and
                  Karel Lenc and
                  Arthur Mensch and
                  Katherine Millican and
                  Malcolm Reynolds and
                  others},
  booktitle = {Proceedings of the International Conference on Neural Information Processing Systems},
  pages   = {23716--23736},
  year    = {2022}
}

@inproceedings{32,
  title   = {Multimodal few-shot learning with frozen language models},
  author       = {Maria Tsimpoukelli and
                  Jacob Menick and
                  Serkan Cabi and
                  S. M. Ali Eslami and
                  Oriol Vinyals and
                  Felix Hill},
  booktitle = {Proceedings of the International Conference on Neural Information Processing Systems},
  pages   = {200--212},
  year    = {2021}
}

@InProceedings{33,
  title        = {Unifying {V}ision-and-{L}anguage tasks via text generation},
  author       = {Jaemin Cho and
                  Jie Lei and
                  Hao Tan and
                  Mohit Bansal},
  booktitle    = {Proceedings of the International Conference on Machine Learning},
  pages        = {1931--1942},
  year         = {2021}
}

@InProceedings{34,
  title        = {Enabling multimodal generation on clip via vision-language knowledge distillation},
  author       = {Wenliang Dai and
                  Lu Hou and
                  Lifeng Shang and
                  Xin Jiang and
                  Qun Liu and
                  Pascale Fung},
  booktitle    ={Findings of the Annual Meeting of the Association for Computational Linguistics},
  pages        = {2383--2395},
  year         = {2022}
}

@inproceedings{35,
  author       = {Liangke Gui and
                  Borui Wang and
                  Qiuyuan Huang and
                  Alexander Hauptmann and
                  Yonatan Bisk and
                  Jianfeng Gao},
  title        = {{KAT:} {A} Knowledge Augmented Transformer for Vision-and-Language},
  booktitle    = {Proceedings of the Conference of the North American Chapter of
                  the Association for Computational Linguistics: Human Language Technologies},
  pages        = {956--968},
  year         = {2022}
}

@inproceedings{36,
  author       = {Yuanze Lin and
                  Yujia Xie and
                  Dongdong Chen and
                  Yichong Xu and
                  Chenguang Zhu and
                  Lu Yuan},
  title        = {{REVIVE:} Regional Visual Representation Matters in Knowledge-Based
                  Visual Question Answering},
  booktitle    = {Proceedings of the International Conference on Neural Information Processing Systems},
pages = {10560--10571},
  year         = {2022}
}

@article{37,
  author       = {Long Chen and
                  Yuhang Zheng and
                  Yulei Niu and
                  Hanwang Zhang and
                  Jun Xiao},
  title        = {Counterfactual Samples Synthesizing and Training for Robust Visual
                  Question Answering},
  journal      = {IEEE Transactions on Pattern Analysis and Machine Intelligence},
  volume       = {45},
  number       = {11},
  pages        = {13218--13234},
  year         = {2023}
}

@article{41,
  author       = {Robert Geirhos and
                  J{\"{o}}rn{-}Henrik Jacobsen and
                  Claudio Michaelis and
                  Richard S. Zemel and
                  Wieland Brendel and
                  Matthias Bethge and
                  Felix A. Wichmann},
  title        = {Shortcut learning in deep neural networks},
  journal      = {Nature Machine Intelligence},
  volume       = {2},
  number       = {11},
  pages        = {665--673},
  year         = {2020}
}

@inproceedings{42,
  author       = {Peize Li and
                  Qingyi Si and
                  Peng Fu and
                  Zheng Lin and
                  Yan Wang},
  title        = {Object Attribute Matters in Visual Question Answering},
  booktitle    = {Proceedings of the {AAAI} Conference on Artificial Intelligence},
  pages        = {18545--18553},
  year         = {2024}
}

@inproceedings{43,
  author       = {Xi Zhang and
                  Feifei Zhang and
                  Changsheng Xu},
  title        = {{VQACL:} {A} Novel Visual Question Answering Continual Learning Setting},
  booktitle    = {Proceedings of the {IEEE/CVF} Conference on Computer Vision and Pattern Recognition},
  pages        = {19102--19112},
  year         = {2023}
}

@inproceedings{44,
  author       = {Haotian Liu and
                  Chunyuan Li and
                  Yuheng Li and
                  Yong Jae Lee},
  title        = {Improved Visual Instruction Tuning},
  booktitle    = {Proceedings of the {IEEE/CVF} Conference on Computer Vision and Pattern Recognition},
pages = {26296--26306},
  year         = {2024}
}

@article{45,
  author       = {Hugo Touvron and
                  Louis Martin and
                  Kevin Stone and
                  Peter Albert and
                  Amjad Almahairi and
                  Yasmine Babaei and
                  Nikolay Bashlykov and
                  Soumya Batra and
                  Prajjwal Bhargava and
                  Shruti Bhosale and
                  others},
  title        = {Llama 2: Open Foundation and Fine-Tuned Chat Models},
  journal={arXiv preprint arXiv:2307.09288},
  year={2023}
}

@inproceedings{46,
  author       = {Teven Le Scao and
                  Thomas Wang and
                  Daniel Hesslow and
                  Stas Bekman and
                  M. Saiful Bari and
                  Stella Biderman and
                  Hady Elsahar and
                  Niklas Muennighoff and
                  Jason Phang and
                  Ofir Press and
                  others},
  title        = {What Language Model to Train if You Have One Million {GPU} Hours?},
  booktitle    = {Findings of the Conference on Empirical Methods in Natural Language Processing},
  pages        = {765--782},
  year         = {2022}
}

@Misc{47,
  title  = {GPT-J-6B: A 6 billion parameter autoregressive language model},
  author = {Ben Wang and Aran Komatsuzaki},
  year   = {2021},
  note = {\url{https://github.com/kingoflolz/mesh-transformer-jax}}
}

@Article{48,
  title   = {Gpt-neo: Large scale autoregressive language modeling with mesh-tensorflow},
  author  = {Sid Black and Leo Gao and Phil Wang and Connor Leahy and Stella Biderman},
  year    = {2021},
  note = {\url{https://doi.org/10.5281/zenodo.5297715}}
}

@inproceedings{61,
  author       = {Xian Zhong and
                  Zipeng Li and
                  Shuqin Chen and
                  Kui Jiang and
                  Chen Chen and
                  Mang Ye},
  title        = {Refined Semantic Enhancement towards Frequency Diffusion for Video
                  Captioning},
  booktitle    = {Proceedings of the {AAAI} Conference on Artificial Intelligence},
  pages        = {3724--3732},
  year         = {2023}
}

@inproceedings{62,
  author       = {Li Yang and
                  Zhiding Xiao and
                  Wenxin Huang and
                  Xian Zhong},
  title        = {StoryLLaVA: Enhancing Visual Storytelling with Multi-Modal Large Language
                  Models},
  booktitle    = {Proceedings of the International Conference on Computational
                  Linguistics},
  pages        = {3936--3951},
  year         = {2025}
}

\end{document}